\pdfoutput=1

\documentclass[11pt]{article} 

\usepackage{EMNLP2022}

\usepackage{times}
\usepackage{latexsym}

\usepackage[utf8]{inputenc}

\usepackage{microtype}

\usepackage{inconsolata}

\usepackage{graphicx}
\usepackage{xcolor}
\usepackage{comment}
\usepackage{booktabs}
\usepackage{amsmath}
\usepackage{amssymb}
\usepackage{amsthm}
\usepackage{bbm} 
\usepackage{subcaption}
\usepackage{multirow}
\usepackage{comment}
\usepackage{makecell} 
\usepackage{tabularx}
\usepackage[normalem]{ulem}
\useunder{\uline}{\ul}{}
\usepackage{hyperref}

\usepackage[export]{adjustbox}
\usepackage{paralist}

\title{Towards No.1 in CLUE Semantic Matching Challenge: Pre-trained Language Model Erlangshen with Propensity-Corrected Loss}

\author{
Junjie Wang$^{1,2}$\footnotemark[1],
Yuxiang Zhang$^{1,2}$\footnotemark[1], 
Ping Yang$^1$, 
Ruyi Gan$^1$, \\
1. CCNL, IDEA, Shenzhen, China\\
2. Waseda University, Tokyo, Japan \\
{\tt wjj1020181822@toki.waseda.jp} \\
{\tt joel0495@asagi.waseda.jp} \\
{\tt \{yangping, ganruyi\}@idea.edu.cn}
}

\begin{document}

\maketitle

\renewcommand{\thefootnote}{\fnsymbol{footnote}}
\footnotetext[1]{This work was performed with equal contribution when they were visiting International Digital Economy Academy (IDEA) as interns.}
\renewcommand{\thefootnote}{\arabic{footnote}}

\begin{abstract}
This report describes a pre-trained language model Erlangshen with propensity-corrected loss, the \textbf{No.1} in \textit{CLUE Semantic Matching Challenge}\footnote{\scriptsize \url{https://www.cluebenchmarks.com/sim.html}}. 
In the pre-training stage, we construct a dynamic masking strategy based on knowledge in Masked Language Modeling (MLM) with whole word masking. 
Furthermore, by observing the specific structure of the dataset, the pre-trained Erlangshen applies propensity-corrected loss (PCL) in the fine-tuning phase. 
Overall, we achieve $72.54$ points in F1 Score and $78.90$ points in Accuracy on the test set. 
Our code is publicly available at:~\url{https://github.com/IDEA-CCNL/Fengshenbang-LM/tree/hf-ds/fengshen/examples/clue_sim}.

\end{abstract}

\section{Background}
\label{sec:background}
\subsection{Semantic Matching}

With the development of data science applications, semantic matching plays a fundamental role in Natural Language Processing (NLP). 
These applications, such as information retrieval, often require semantic matching to combine data from multiple sources prior to further analysis~\cite{li2021deep/semantic_matching_survey}. 
It is natural to think of semantic matching as a binary classification issue. 
Given a pair of sentences, the artificial intelligent system is required to determine if two sentences provide the same meaning.

\subsection{CLUE Semantic Matching Challenge}

To up the challenge of this problem, the CLUE team~\cite{xu-etal-2020-clue/clue} releases \textit{CLUE Semantic Matching Challenge}. 
In detail, the QBQTC (QQ Browser Query Title Corpus) challenge dataset includes three labels, which are \{$0$, irrelevant or not matching; $1$, some correlation; $2$, high relevant\}, instead of only two. 
Furthermore, QBQTC dataset is designed as a learning-to-rank (LTR) dataset that integrates relevance, authority, content quality, timeliness and other dimensional annotations, which is widely used in QQ search engine business scenarios\footnote{\scriptsize \url{https://browser.qq.com/}}. 
In additional, this challenge is included in CLUE benchmark, which is widely evaluated in Chinese NLP community.

\section{Introduction}
\label{sec:introduction}

In Chines Natural Language Understanding (NLU) tasks, Pre-trained Language Models (PLMs) have proven to be effective, such as BERT~\cite{DBLP:conf/naacl/DevlinCLT19/bert}, MacBERT~\cite{DBLP:conf/emnlp/CuiC000H20/macbert} and ALBERT~\cite{DBLP:conf/iclr/LanCGGSS20/albert}. 
MacBERT and RoBERTa improve the pre-training tasks of BERT and ALBERT by considering Chinese grammatical structure~\cite{DBLP:conf/emnlp/CuiC000H20/macbert}. 
However, those models only randomly masks tokens, which leads to a simple pre-training task. 
Therefore, we introduce Knowledge-based Dynamic Masking (KDM) method to mask semantically rich tokens. 
Moreover, we pre-train Erlangshen in a large-scale Chinese corpus and optimize the pre-training settings such as employing pre-layer normalization~\cite{DBLP:conf/icml/XiongYHZZXZLWL20/pre-ln} method. 
In addition, several designed prompts are assigned into surprised datasets for processing surprised pre-training tasks.

As a specific Chinese NLU task, the \textit{CLUE Semantic Matching Challenge} has higher difficulty than binary classification semantic matching. 
By our observation, the QBQTC dataset has serious data imbalance problem, which includes a imbalance label distribution with severe bias. 
The large number of negative samples drives the model to ignore learning difficult positive samples under the effect of popular cross-entropy. 
Therefore, the poor prediction on positive samples directly leads to a low F1 score. 
This problem comes from the nature of cross-entropy, which treats each sample equally and tries to push them to positive or negative. 
An workable loss to address unbalanced samples is  Dice loss~\cite{DBLP:conf/acl/LiSMLWL20/dice}. 
In classification scenario, we only need to concern a probability $< 0.5$ instead of an extract probability $0$ when it is a negative sample. 
Although Dice loss achieves the success of data imbalance, it only consider target categories while ignoring non-target categories. 
Therefore, this method is susceptible to the influence of a single sample. 
Inspired by Label-smoothing~\cite{DBLP:conf/cvpr/SzegedyVISW16/labelsmooth}, we propose Propensity-Corrected Loss (PCL) to correct the propensity of the system to predict the labels.

Through extensive experiments on the benchmark, we demonstrate that our approach achieve state-of-the-art (SOTA) performance. 
Ablation studies reveal that Erlangshen model and PCL are the key components to the success of understanding semantic matching. 
Our contributions are as follows.

\begin{itemize}
\setlength{\itemsep}{1pt}
\setlength{\parskip}{0pt}
\setlength{\parsep}{0pt}
    \item We apply a powerful PLM Erlangshen on a Chinese semantic matching task.
    \item Our proposed PCL teaches the model well in an imbalance dataset.
    \item Our approach win the No.1 over the existing SOTA models on \textit{CLUE Semantic Matching Challenge}. 
\end{itemize}




\section{Approach}
\label{sec:approach}

In this section, we introduce the Pre-trained Language Model (PLM) Erlangshen (Section~\ref{sec:erlangshen}) and a new Propensity-Corrected Loss (PCL) (Section~\ref{sec:pcl}) to address a semantic matching problem.

\subsection{PLM: Erlangshen}
\label{sec:erlangshen}

To improve BERT~\cite{DBLP:conf/naacl/DevlinCLT19/bert} in Chinese tasks, we release a new PLM, Erlangshen. 
Note that we \textbf{can not} explain the details of Erlangshen, such as training implementation and detailed methods. 
This advanced Chinese PLM follows the private policy and company agreement of IDEA\footnote{\scriptsize \url{https://idea.edu.cn/}}. 
Fortunately, we are allowed to describe the idea of the design and the basic structure. 
Although Erlangshen is kept under wraps, we open-source its weights in Huggingface\footnote{\scriptsize \url{https://huggingface.co/IDEA-CCNL}} for accessible use in custom tasks.

After the necessary declarations, we introduce our basic ideas for constructing Erlangshen. 
Our Erlangshen employs the similar multi-layer transformer architecture as BERT. 
By following the pre-training and fine-tuning paradigm, Erlangshen is pre-trained with Masked Language Modeling (MLM)~\cite{DBLP:conf/naacl/DevlinCLT19/bert} and Sentence-Order Prediction (SOP)~\cite{9599397/bert_wwm} task. 
When considering the linguistic characteristics of Chinese, we apply Whole Word Masking (WWM)~\cite{9599397/bert_wwm} to process whole Chinese words instead of individual Chinese characters in WordPiece. 
Furthermore, since the original MLM strategy only randomly masks tokens, the language model can easily recover common tokens such as ``I am''. 
To address this issue, we propose a Knowledge-based Dynamic Masking (KDM) method to mask semantically rich tokens such as entities and events. 
Besides understanding intra-sentence information, we apply SOP task to learn inter-sentence information instead of next sentence prediction task. 
The main reason is that SOP is much more effective than NSP~\cite{9599397/bert_wwm, DBLP:conf/iclr/LanCGGSS20/albert}.
After pre-training on large-scale unsupervised data, Erlangshen and its variants are released in Huggingface.

In fine-tuning phase, we add a classifier layer on the top of Erlangshen and predict its label. 
In detail, given a pair of sentences $S_1$ and $S_2$, we generate the input ($x_{inp}$) with special tokens in BERT~\cite{DBLP:conf/naacl/DevlinCLT19/bert} as follows:

\begin{equation}
x_{inp} = {\tt [CLS]} S_1 {\tt [SEP]} S_2 {\tt [SEP]}
\end{equation}

Then, we fed {\tt [CLS]} as the global feature into the classifier for generating the final answer.

\subsection{Propensity-Corrected Loss}
\label{sec:pcl}

By observing the QBQTC dataset with three labels, it suffer data imbalance problem (Details in Section~\ref{sec:data_analysis}). 
Since the label $1$ occupies the majority of samples, we focus on the correlation between different categories. 
We introduce the Propensity-Corrected Loss (PCL) to encourage the model to predict the dominant categories, which corrects the predicting propensity.

We consider a label set \{$0$, irrelevant or not matching; $1$, some correlation; $2$, high relevant\}. 
Given a pair of sentences ($S_1$ and $S_2$), we argue that the model must take care of the ``distance'' of labels. 
For example, if it is a not-matching pair, the penalty for the model to misclassify it as label $2$ must be larger than as label $1$. 
Because label $1$ is closer to label $0$ in ``distance'' than label $2$.
If the opposite occurs, a loss bonus (which reduces the original loss) is provided. 
Similarly, this case also works with misclassifying labels $1$ into labels $0$ or $2$. 
Note that the label $0$ represents irrelevant; the penalty for assigning the label $1$ to the label $0$ is also greater than assigning it to the label $2$.

In detail, considering simple implementation, we apply PCL based on all samples. 
Let $label \in \{ label_0, label_1, label_2 \}$ to be the ground truth set. 
Since PCL considers multiple conditions, we compute $PCL_{+}$ and $PCL_{-}$ as follow.

\begin{equation}
\begin{small}
PCL_{+} = \varepsilon \frac{\sum_{i=1}^{N}Label_0^i}{\sum_{i=1}^{N}Label_1^i + \sum_{i=1}^{N}Label_2^i},
\end{small}
\end{equation}

\begin{equation}
\begin{small}
PCL_{-} = \varepsilon \frac{\sum_{i=1}^{N}Label_0^i + \sum_{i=1}^{N}Label_2^i}{\sum_{i=1}^{N}Label_1^i},
\end{small}
\end{equation}
where $N$ is the sample number of the dataset and $\varepsilon$ is a scaling ratio.

Then, the PCL ($\mathcal{L}$) can be calculated as:
\begin{equation}
\mathcal{L}=\left\{\begin{matrix} 
LS - PCL_{-},  & \text{Condition 1}, \\
LS + PCL_{+},  & \text{Condition 2}, \\
LS,  & \text{Condition 3},
\end{matrix}\right.
\end{equation}
where $LS$ represents the Label-smoothing loss. 
\text{Condition 1} tends to encourage the model, e.g., the model misclassify label $1$ into label $2$, which is better than label $0$. 
\text{Condition 2} tends to punish the model, e.g., the model misclassifies label $2$ into label $0$. 
\text{Condition 3} considers the cases where no correction propensity is required.

\subsection{Discussion}

Interestingly, PCL can be computed within a batch rather than across all samples. 
Therefore, as a future work, we will consider PCL in local level and global level.

\section{Experiment}
\label{sec:experiment}
\subsection{Experimental Setup}

\label{sec:data_analysis}

We summarize QBQTC dataset splits in Table~\ref{table:dataset_division}. 
We further investigate the average number of words for $S_1$ and $S_2$, average token number after organization and the label ratio. 
The dataset details are shown in Table~\ref{table:dataset_detail}.

\begin{table}[!tb]
\begin{center}
\begin{small}
\begin{adjustbox}{max width=0.48\textwidth}
\begin{tabular}{cccc}
\toprule
\textbf{Train} & \textbf{Dev} & \textbf{Test\_public} & \textbf{Test}                   \\ 
\midrule
180,000 & 20,000 & 5,000         & $\geq 100,000$ \\
\bottomrule
\end{tabular}
\end{adjustbox}
\end{small}
\end{center}
\caption{Statistics of the QBQTC dataset splits.}
\label{table:dataset_division}
\end{table}

\begin{table}[!tb]
\begin{center}
\begin{small}
\begin{adjustbox}{max width=0.48\textwidth}
\begin{tabular}{lccc}
\toprule
 & \textbf{Train} & \textbf{Dev} & \textbf{Test\_public}  \\ 
\midrule
Avg. $S_1$ & 9.6 & 8.60 & 9.7        \\
Avg. $S_2$ & 25.4 & 25.6 & 25.3        \\
Avg. Token & 33.9 & 34.1 & 33.8        \\
Label Ratio & 2~:~5~:~1 & 2~:~5~:~1 & 2~:~5~:~1        \\
\bottomrule
\end{tabular}
\end{adjustbox}
\end{small}
\end{center}
\caption{QBQTC dataset details.}
\label{table:dataset_detail}
\end{table}

\begin{table*}[!tb]
\begin{center}
\begin{normalsize}
\begin{adjustbox}{max width=\textwidth}
\begin{tabular}{llcc}
\toprule
\textbf{Model}                 & \textbf{Research Institute} & \textbf{F1 Score} & \textbf{Accuracy} \\ \midrule
Erlangshen            & IDEA               & \textbf{72.54}   & \textbf{78.94}    \\
DML RoBERTa           & CMB AI Lab         & 72.52   & 78.25    \\ 
roberta-large         & CaiBao             & 72.14   & 78.54    \\
bert                  & daydayup           & 71.43   & 77.38    \\
T5                    & BoTong             & 67.07   & 74.63    \\
first\_model          & nlp\_tech          & 66.77   & 74.05    \\
RoBERTa-wwm-ext-large & CLUE               & 66.30   & 73.10     \\
BoTong                & BoTong             & 65.60   & 74.17    \\
Bert-base-chinese     & CLUE               & 64.10   & 71.80     \\
RoBERTa-wwm-ext       & CLUE               & 64.00   & 71.00      \\ \bottomrule
\end{tabular}
\end{adjustbox}
\end{normalsize}
\end{center}
\caption{Leaderboard of CLUE Semantic Matching Challenge. The results are reported by CLUE team and the best performance is in \textbf{bold}. Besides, we have converted several Chinese terms into English for easy description and the original one presents in Figure~\ref{fig:screenshot}. Since there is no specific meaning, it will not affect the understanding of the leaderboard and this paper.}
\label{table:main_result}
\end{table*}


In all our experiments, we apply the variant ``Erlangshen-MegatronBert-1.3B-Similarity''\footnote{\scriptsize \url{https://huggingface.co/IDEA-CCNL/Erlangshen-MegatronBert-1.3B-Similarity}} as ``Erlangshen''. 
The detailed settings are followed the open-sourced code\footnote{\scriptsize \url{https://github.com/IDEA-CCNL/Fengshenbang-LM/tree/hf-ds/fengshen/examples/clue_sim}}. 
We tune the models in train and dev set, and evaluate them in test\_public set for easy analysis. 

\subsection{Challenge Results}

After submitting our prediction files to the CLUE official website, the comparison with existing SOTA performance is listed in Table~\ref{table:main_result}. 
Our proposed Erlangshen with PCL outperforms all models, which achieves $72.54$ points in F1 Score and $78.90$ points in Accuracy.

\begin{table}[!tb]
\begin{center}
\begin{small}
\begin{adjustbox}{max width=0.48\textwidth}
\begin{tabular}{lcccc}
\toprule
           & \multicolumn{2}{c}{Label smoothing} & \multicolumn{2}{c}{PCL}                                                                                                             \\ \cmidrule{2-5} 
           & F1 Score          & Accuracy        & F1 Score                                                         & Accuracy                                                         \\ \midrule
RoBERTa-wwm   & 67.94             & 75.02            & \begin{tabular}[c]{@{}c@{}}70.03\\ (+2.09)\end{tabular}          & \begin{tabular}[c]{@{}c@{}}77.02\\ (+2.00)\end{tabular}          \\
MacBERT    & 69.18             & 76.58            & \begin{tabular}[c]{@{}c@{}}71.78\\ (+2.60)\end{tabular}          & \begin{tabular}[c]{@{}c@{}}78.38\\ (+1.80)\end{tabular}          \\
Erlangshen & \textbf{72.24}    & \textbf{79.10}   & \textbf{\begin{tabular}[c]{@{}c@{}}72.74\\ (+0.50)\end{tabular}} & \textbf{\begin{tabular}[c]{@{}c@{}}80.00\\ (+0.90)\end{tabular}} \\ \bottomrule
\end{tabular}
\end{adjustbox}
\end{small}
\end{center}
\caption{Ablation experiments with normal Label-smoothing Loss and Propensity-Corrected Loss on Test\_public set. The best results are in \textbf{bold}. ``RoBERTa-wwm''\protect\footnotemark[8] and ``MacBERT''\protect\footnotemark[9]~\cite{DBLP:conf/emnlp/CuiC000H20/macbert} are implemented by accessing Huggingface.}
\label{table:ablation}
\end{table}

\subsection{Ablation Studies}

To verify the effectiveness of our Erlangshen and Propensity-Corrected Loss (PCL), we conduct ablation studies on Test\_public set, as shown in Table~\ref{table:ablation}.

We first verified the effectiveness of our PLM Erlangshen.
Regardless of the loss employed, our model outperform widely used MacBERT and RoBERTa-wwm. 
For label smoothing loss, our Erlangshen improve $3.06$ points on F1 Score and $2.52$ points on Accuracy over MacBERT.

Then, we explore the effectiveness of the proposed PCL. 
Among the models, PCL helps all models to achieve a better performance than the original label smoothing one. 
For MacBERT, PCL can provide up to $2.60$ points improvement in F1 score, and up to $1.80$ points improvement in accuracy. 
For Erlangshen, PCL can benefit up to $2.61$ points improvement in F1 score, and up to $0.90$ points improvement in accuracy.
However, Erlangshen obtains less gain from PCL than other models. 
Our powerful Erlangen might already learn imbalance information in the dataset, so it has learned some propensity by itself rather than loss' help.

\footnotetext[8]{\scriptsize \url{https://huggingface.co/hfl/chinese-roberta-wwm-ext-large}}

\footnotetext[9]{\scriptsize \url{https://huggingface.co/hfl/chinese-macbert-large}}

\section{Conclusion}
\label{sec:conclusion}
In this work, we introduce Erlangshen with novel KDM method in MLM. 
Erlangshen also benefit from large-scale unlabeled data and prompted surprised datasets. 
By observing QBQTC dataset, we propose a new Propensity-Corrected Loss to encourage the model to predict the dominant categories, which corrects the predicting propensity. 
Our approach achieve the best results on \textit{CLUE Semantic Matching Challenge} leaderboard. 
The ablation studies demonstrate the effectiveness of our designs.

In future, we plan to explore the Propensity-Corrected Loss in local and global level of the dataset. 
On the other hand, we will investigate the capability of Erlangshen in other NLP tasks.

\section*{Ethical Consideration}
\label{sec:ethical}
Natural language processing is an important technology in our society. 
It is necessary to discuss its ethical influence~\cite{DBLP:conf/ethnlp/LeidnerP17/ethical_desgin}. 
In this work, we develop a novel language model and the PCL loss. 
As discussed in~\cite{schramowski2022large/ethic_large_model, DBLP:journals/corr/abs-1912-05238/ethic_bert,DBLP:conf/acl/BlodgettBDW20/ethic_language}, language models might contain human-like biases from pre-trained datasets, which might embed in both the parameters of the models and outputs. 
In additional, we know that Erlangshen with PCL loss can be employed in similar applications, which may have potential to be abused and caused unexpected influence. 
We encourage open debating on its utilization, such as the task selection and the deployment, hoping to reduce the chance of any misconduct.

\begin{figure*}[]
    \centering
    \includegraphics[width=\textwidth]{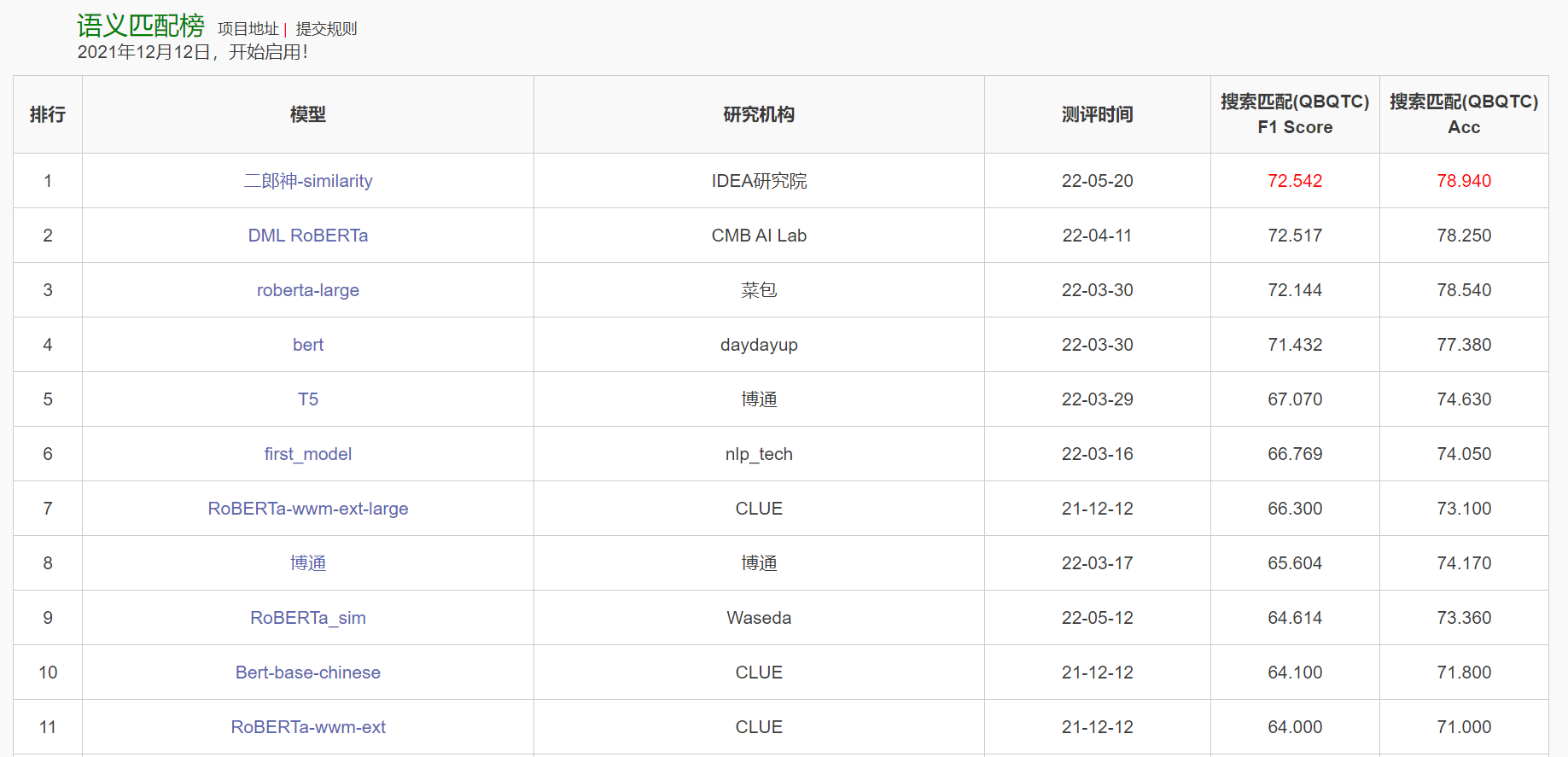}
    \caption{A screenshot of CLUE Semantic Matching Challenge official website. Access Date: July 10, 2022.}
    \label{fig:screenshot}
\end{figure*}

\newpage

\bibliography{anthology, custom}
\bibliographystyle{acl_natbib}


\appendix

\end{document}